\pdfoutput=1

\documentclass[11pt]{article}

\usepackage[]{acl}

\usepackage{times}
\usepackage{latexsym}

\usepackage[T1]{fontenc}

\usepackage[utf8]{inputenc}

\usepackage{microtype}

\usepackage[utf8]{inputenc}
\usepackage{inconsolata}
\usepackage{graphicx}
\usepackage{amsmath,amsfonts,amssymb}
\usepackage{siunitx}
\usepackage{xspace}
\usepackage{multicol}
\usepackage{multirow}
\usepackage{rotating}
\usepackage{booktabs}
\usepackage{cleveref}
\usepackage{hyperref}
\usepackage{subcaption}

\newcommand{\monarch}{Monarch\xspace}

\title{An Empirical Investigation of Matrix Factorization Methods for Pre-trained Transformers}

\author{Ashim Gupta, Sina Mahdipour Saravani,  P. Sadayappan, Vivek Srikumar \\
        Kahlert School of Computing \\ University of Utah  \\ \texttt{ashim@cs.utah.edu}}

\begin{document}
\maketitle
\begin{abstract}

The increasing size of transformer-based models in NLP makes the question of compressing them important. In this work, we present a comprehensive analysis of factorization based model compression techniques. Specifically, we focus on comparing straightforward low-rank factorization against the recently introduced Monarch factorization~\cite{dao2022monarch}, which exhibits impressive performance preservation on the GLUE benchmark.
%
%
%
To mitigate stability issues associated with low-rank factorization of the matrices in pre-trained transformers, we introduce a staged factorization approach wherein layers are factorized one by one instead of being factorized simultaneously. Through this strategy we significantly enhance the stability and reliability of the compression process. Further, we introduce a simple block-wise low-rank factorization method, which has a close relationship to Monarch factorization. 
Our experiments lead to the surprising conclusion that straightforward low-rank factorization consistently outperforms Monarch factorization across both different compression ratios and six different text classification tasks.
\end{abstract}

\section{Introduction}

Massive pretrained transformers excel at diverse language understanding tasks. For example, on the GLUE benchmark\footnote{\url{https://gluebenchmark.com/leaderboard}}, 22 systems achieve superhuman scores, with the best one containing 4.6B parameters and a 4+ point improvement. Today's largest models have hundreds of billions of parameters~\cite{brown2020language}, and are growing.
However, their size\,---\,and associated substantial GPU memory and disk space requirements\,---\,hinders deployment on low-end devices. This prompts the need for effective model compression. 

Common strategies include quantization for reduced numerical precision~\citep{wu2016quantized,hubara2018quantized,stock2019and}, network pruning for removing redundant components~\citep{lecun1989optimal,han2015learning,michel2019sixteen,fan2019reducing}, and distillation, involving training a smaller student network supervised by a larger teacher network~\citep{hinton2015distilling,sanh2019distilbert,jiao2020tinybert}. 
A relatively under-explored approach involves factorizing parameter matrices into smaller ones; we focus on exploring methods in this category. 

A natural low-rank factorization approach represents a large matrix as the product of two smaller matrices, thereby lowering rank by construction. 
%
Borrowing the popular strategy of fine-tuning pretrained models, dense pretrained weight matrices can be approximated using singular value decomposition (SVD)~\citep{sainath2013low,idelbayev2020low}, and then fine-tuned for a task.
\citet{dao2022monarch} recently proposed an alternative called \monarch factorization, which represents a matrix with a product of two block-diagonal matrices, mediated by element permutations. This approach leverages a variant of SVD to approximate the two block-diagonal factors from a pretrained dense matrix. We seek to systematically compare such factorization methods.

Any method for factorizing a model  requires first projecting (i.e. approximating) its pretrained weight matrices into the factorized forms. 
We find that na\"ively doing so can cause stability issues for downstream fine-tuning: many factorized models do not train at all. We call these \textit{failed runs}.\footnote{The issue of failed runs for un-factorized pre-trained transformers is highlighted in~\citet{dodge2020fine}} Compared to the original model, we observe a much higher proportion of failed runs in the factorized models (especially with low-rank factorization). We propose a simple, layer-wise factorization strategy to alleviate these issues.

We conduct a comprehensive set of experiments on 6 tasks from the GLUE benchmark and with 4 different pre-trained models. Our experiments reveal that the simple, easy-to-implement low-rank factorization consistently outperforms more sophisticated strategies, including  \monarch factorization. 

The insights from our work are:

\begin{itemize}
  \item \textbf{Stability of Factorized Models.} Our findings reveal notable stability issues in factorized transformer models, particularly with low-rank factorization. To address this, we introduce a straightforward training strategy called Staged Factorization. Instead of factorizing the entire model in one step, we compress the model in stages. The results demonstrate a significant improvement in stability for low-rank factorization, bringing it close to the stability level observed in the unfactorized model.  
  \item \textbf{Dominance of Low-Rank Factorization.} Extensive experimentation on six classification datasets from the GLUE benchmark establishes Low-Rank Factorization as the superior choice to a more sophisticated, recently proposed Monarch Factorization. This factorization method not only achieves high accuracy but also exhibits the lowest inference times. Additionally, it proves to be straightforward to implement, making it a compelling option for transformer models.
\end{itemize}
\section{Matrix Factorizations}
\label{sec:factors}
\begin{figure}
\includegraphics[trim={0mm 0mm 0cm 0cm},clip, width=\linewidth]{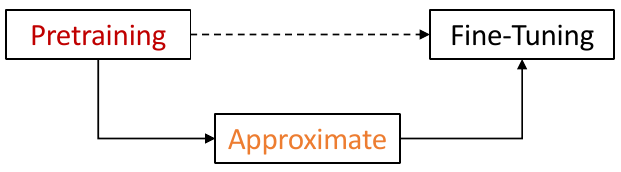}
\caption{ The dotted line represents traditional fine-tuning for the non-factorized model initialized with pretrained weights. The solid line signifies an intermediate step involving the projection of pretrained weights onto factorized matrices, followed by task-specific fine-tuning.
}
\label{fig:approximate}
\end{figure}
Task-specific fine-tuning typically commences by initializing the model from a pretrained checkpoint, such as BERT~\citep{devlin2019bert}. 
If we seek to build many factorized models of different sizes, it is impractical to pre-train each one from scratch; both cost and time are prohibitive.
%
Instead, a natural strategy is to proceed via an intermediate step of projecting pre-trained weights into a factorized form, followed by the downstream task-specific fine-tuning. We term this approach the \textit{approximate then fine-tune} approach (~\cref{fig:approximate}). 
Note that, the approximation step is relatively cost-effective compared to the resource-intensive pre-training. Note that, since matrices in existing pretrained models are close to full-rank (see ~\cref{fig:spectrum} in the Appendix), the projection step provides only an approximation to the original parameters.

Transformer architectures concentrate a significant portion of their learnable parameters within the embedding matrix, the four linear layers of each self-attention block~\footnote{These are query, key, value, and the output matrices.} and two linear layers in each feed-forward network.
We factorize all parameter layers except the embedding matrix, since during preliminary experiments we observed that the embedding matrix exhibits greater sensitivity to factorization.


We consider each dense pre-trained matrix $\mathbf{W} \in \mathbb{R}^{m \times n}$ that needs to be factorized into a smaller set of matrices. 
With $mn$ parameters, achieving compression entails a factorization that reduces the parameter count below $mn$.

\subsection{Low-Rank Factorization}
The simplest way to compress this potentially large matrix is to factorize it into a product of two smaller matrices, $\mathbf{W} = \mathbf{U} \mathbf{V}^\top$, where $\mathbf{U} \in \mathbb{R}^{m \times r}$, $\mathbf{V} \in \mathbb{R}^{n \times r}$, and $r$ is the rank of the factorization. This requires $r(m+n)$ trainable parameters, and compression is achieved when $r < \frac{mn}{m+n}$, and is a hyperparameter that is dependent on the desired compression level. 

For a given rank $r$, the initialization of the weights requires a low-rank approximation which can be efficiently computed using singular value decomposition, and truncating it to the first $r$ singular values. Since pre-trained weight matrices are near full-rank, the accuracy of this approximation depends on the rank, introducing a trade-off between the compression level and approximation quality.



\subsection{Block Low-Rank Factorization}
The objective of low-rank factorization is to capitalize on redundancies or patterns within the parameters present in the weight matrix. We consider an alternative variant called Block Low-Rank Factorization, which seeks to exploit redundancies in a more localized manner, specifically at the level of a sub-block within the parameter matrix. 
We denote this parameterization as a 4D tensor $\mathbf{\mathcal{W}}_{ijkl} \in \mathbb{R}^{b_1 \times b_2 \times o \times p}$, where $i$, $j$ are the two block indices, and $k$, $l$ refer to the row and column indices inside these blocks.\footnote{We write 3D and 4D tensors with indicies to clarify their dimensionality, and also to facilitate their use in the Einstein notation.}
Further, $b_1$, $b_2$ refer to the number of blocks, $o$, $p$ are the dimensionalities of the blocks and satisfy $b_1 = \frac{m}{o}$, and $b_2 = \frac{n}{p}$. By design, we can have the same number of blocks on two axes ($b_1 = b_2 = b$) by making the blocks rectangular. 
A rank $r$ factorization can then be written in the Einstein notation as:
\begin{align*}
\mathbf{\mathcal{W}}_{ijkl} = \mathbf{\mathcal{L}}_{ijkr} \mathbf{\mathcal{R}}_{ijrl} 
\end{align*}
where, $\mathbf{\mathcal{L}}_{ijkr} \in \mathbb{R}^{b_1 \times b_2 \times o \times r}$ and $\mathbf{\mathcal{R}}_{ijrl} \in \mathbb{R}^{b_1 \times b_2 \times r \times p}$. This is straightforward to implement using the \texttt{einsum} functionality available in modern deep learning frameworks. The tensor $\mathbf{\mathcal{L}}_{ijkr}$ contains $b\times b\times o\times r = b m r$ parameters, and $\mathbf{\mathcal{R}}_{ijrl}$ contains $b n r$ parameters, with the factorization containing a total of $br(m+n)$ parameters. 

Notably, setting number of blocks to 1, i.e. $b_1 = b_2 = 1$, and $o = m$, $p = n$ recovers the original low-rank factorization over the full matrix. The initialization from pretrained weights is done by performing SVD on each block independently for a total of $b_1b_2$ SVD's.

\subsection{Monarch Factorization}
~\citet{dao2022monarch} proposed a new factorization called the \monarch Factorization. We briefly describe it here, and refer the reader to the original work for details. \monarch  factorizes a dense matrix $\mathbf{M}$ as a product of two block-diagonal matrices up to permutation:
\begin{align*}
\mathbf{M} = \mathbf{P_1}\mathbf{L}\mathbf{P_2}^\top\mathbf{R}
\end{align*}
Here, $\mathbf{L}$ and $\mathbf{R}$ are the two block-diagonal matrices and $\mathbf{P_1}$, and $\mathbf{P_2}$ are the pre-defined permutation matrices. 
Pre-multiplying with a permutation matrix $\mathbf{P}$ permutes the rows of the matrix, and so $\mathbf{P_1} \mathbf{L}$ and $\mathbf{P_2}^\top \mathbf{R}$ are row-permuted variants of the matrices $\mathbf{L}$ and $\mathbf{R}$. When $\mathbf{M}$ is a square matrix, $\mathbf{P_1} = \mathbf{P_2}$ and $\mathbf{P_2} = \mathbf{P_2}^\top$. Additionally, for a rectangular matrix $\mathbf{M}$, the permutation matrices $\mathbf{P_1}$ and $\mathbf{P_2}$ will have different shapes.

Due to inherent sparsity in block-diagonal matrices, these can be more compactly stored as 3D tensors $\mathbf{\mathcal{L}}_{ikl}$ and $\mathbf{\mathcal{R}}_{ikl}$, where $i$ denotes the index of diagonal block in matrix and $k$, $l$ refer to the indices inside each diagonal block. The original \monarch work admits a rank for the factorization, though \citet{dao2022monarch} only explored rank-1 factorization. With a rank-1 factorization, we have $\mathbf{\mathcal{L}}_{ikl} \in \mathbb{R}^{b \times b \times o}$ and $\mathbf{\mathcal{R}}_{ikl} \in \mathbb{R}^{b \times b \times p}$. For a general rank $r$ factorization, $\mathbf{\mathcal{L}}_{ikl} \in \mathbb{R}^{b \times br \times o}$ and $\mathbf{\mathcal{R}}_{ikl} \in \mathbb{R}^{b \times br \times p}$, where $b$ and $r$ are the number of blocks and the rank respectively. The number of learnable parameters is therefore $b \times br \times o + b \times br \times p = br(m+n)$.


\paragraph{Monarch as a variant of Block Low-Rank}
Notice that, for the same block size $b$ and rank $r$, Block Low-Rank and \monarch factorizations contain the same number of learnable parameters. This is not a coincidence. 
We find that Monarch factorization can be seen as an extension of Block Low-Rank factorization. We show an example in the appendix~\cref{fig:example}.

Intuitively, while Block Low-Rank factorization exploits redundancies of localized blocks of elements of a matrix, the permutation operation in Monarch helps to exploit more distant redundancies by first permuting elements of the original dense matrix followed by a low-rank factorization of blocks of this permuted dense matrix.

One might ask, why consider Block Low-Rank factorization at all? It serves two purposes here: Firstly, as an extension of Low-Rank factorization to smaller, arbitrarily sized blocks, it allows an examination of whether the exploitation of more local redundancies by Block Low-Rank is beneficial for effective factorization. Secondly, since Monarch is an extension of Block Low-Rank, we can investigate whether the more distant block redundancies exploited by Monarch are advantageous. 

\begin{table}[]
\centering
\begin{tabular}{lrrc}
\toprule
\textbf{Dataset} & \textbf{|Train|} & \textbf{|Eval|} & \textbf{Metric} \\ \midrule
MNLI             & 392,702          & 19,647          & Acc.            \\
QNLI             & 104,743          & 5,463           & Acc.            \\
RTE              & 2,490            & 277             & Acc.            \\
SST-2            & 67,349           & 872             & Acc.            \\
QQP              & 363,845          & 40,430          & Acc./F1         \\
MRPC             & 3,668            & 408             & Acc./F1         \\ \bottomrule
\end{tabular}
\caption{Datasets used in our experiments. We use the official splits provided by the GLUE benchmark.}
\label{tab:dataset}
\end{table}

\section{Stability Issues in Factorized Models}
\begin{figure}
\includegraphics[trim={0mm 1mm 0cm 1cm},clip, width=\linewidth]{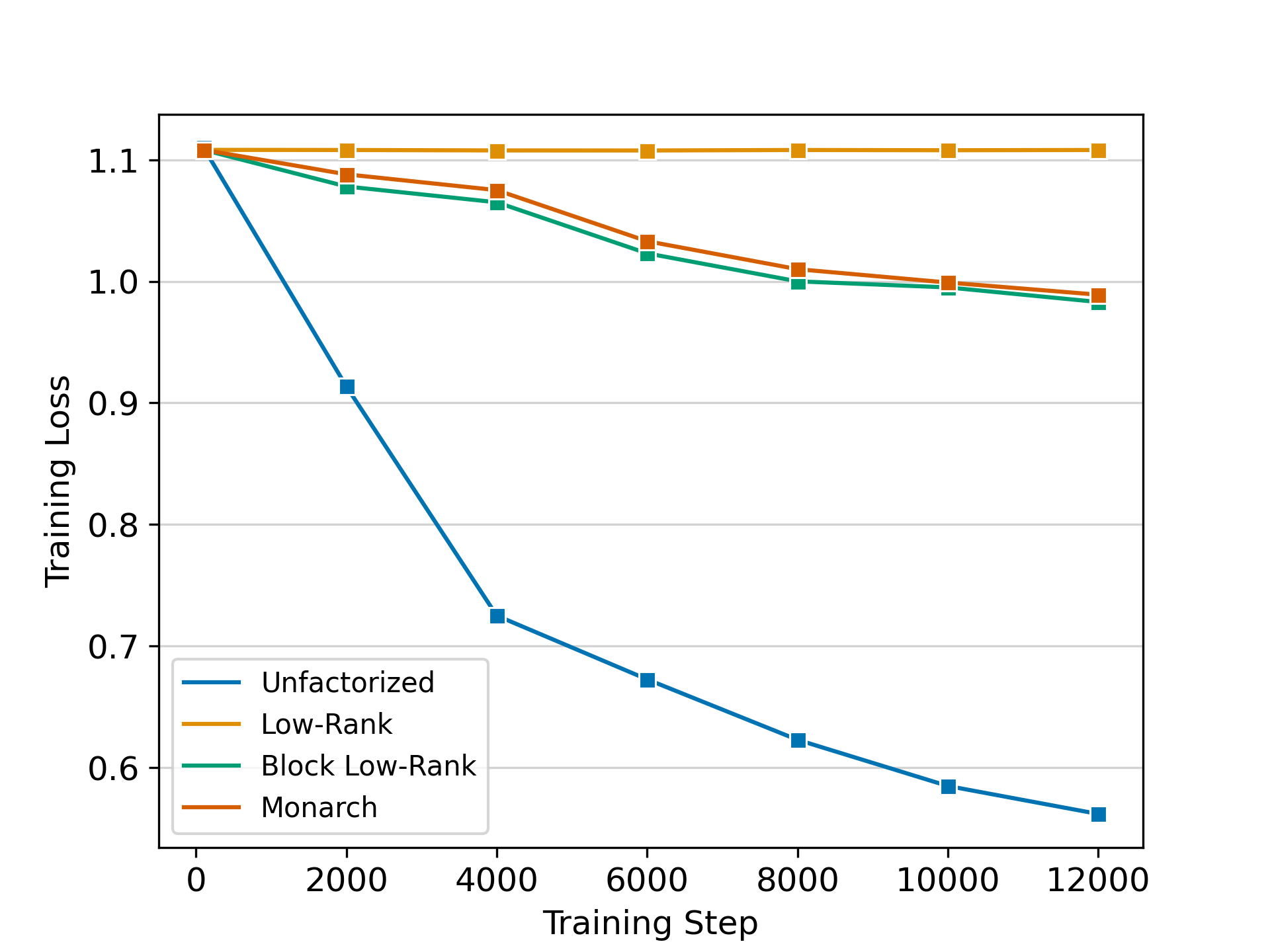}
\caption{ Comparison of the training loss of the factorized and unfactorized models. The loss curves are shown for an unfactorized BERT model vs factorized models (25\% parameters) trained on the MNLI dataset containing three labels with initial learning rate of 2e-5.  
}
\label{fig:loss}
\end{figure}
~\citet{dodge2020fine} showed that supervised fine-tuning with pre-trained models can be brittle and the performance numbers can vary greatly across different random runs of the same model. They refer to this as \textit{fine-tuning instability}. 
More specifically, this instability can occur in two ways: \textit{high standard deviation} across random runs, and \textit{failed runs}. 
Failed runs refer to those runs when the model does not train at all. The classification accuracy in these cases is less or equal to that of the majority classifier on the training dataset. 

While unfactorized models can occasionally suffer from fine-tuning instability, we observe that this occurs more frequently for factorized models and is a bigger issue for such models.~\Cref{fig:loss} shows one such case for a factorized model with 28.3 M parameters (compressed from 109.5 M BERT base model). During our preliminary experiments, we observed that the factorized models are also more sensitive to the choice of initial learning rate, making hyperparameter search for learning rates more important. For these factorized models, increasing training epochs (up to 100) does not help either. 

\paragraph{Staged Factorization}
We propose a simple training strategy to help alleviate some of these issue. A potential factor contributing to the heightened instability is our simultaneous factorization of every weight matrix across all layers. This simultaneous factorization resets the initialization of the entire model, leading to challenges in optimization. This implies an alternative training approach involving the gradual factorization and initialization of layers in multiple stages. Rather than factorizing all matrices simultaneously, we opt to factorize only a subset of matrices at each stage. Following each stage of factorization, the model undergoes a few training steps before progressing to the next phase of factorization.
~\footnote{We use 500 training steps for each stage.} 
There are two possible staging strategies: the \textit{low-to-high} approach, which involves factorizing matrices from the first layer to the last, and the \textit{high-to-low} strategy, which starts factorization from the last layer and progressively includes initial layers later in the process. We find that high-to-low approach provides higher stability and so we choose that for all our subsequent experiments. 

\begin{figure}
\includegraphics[trim={0mm 0.42cm 0cm 0cm},clip, width=\linewidth]{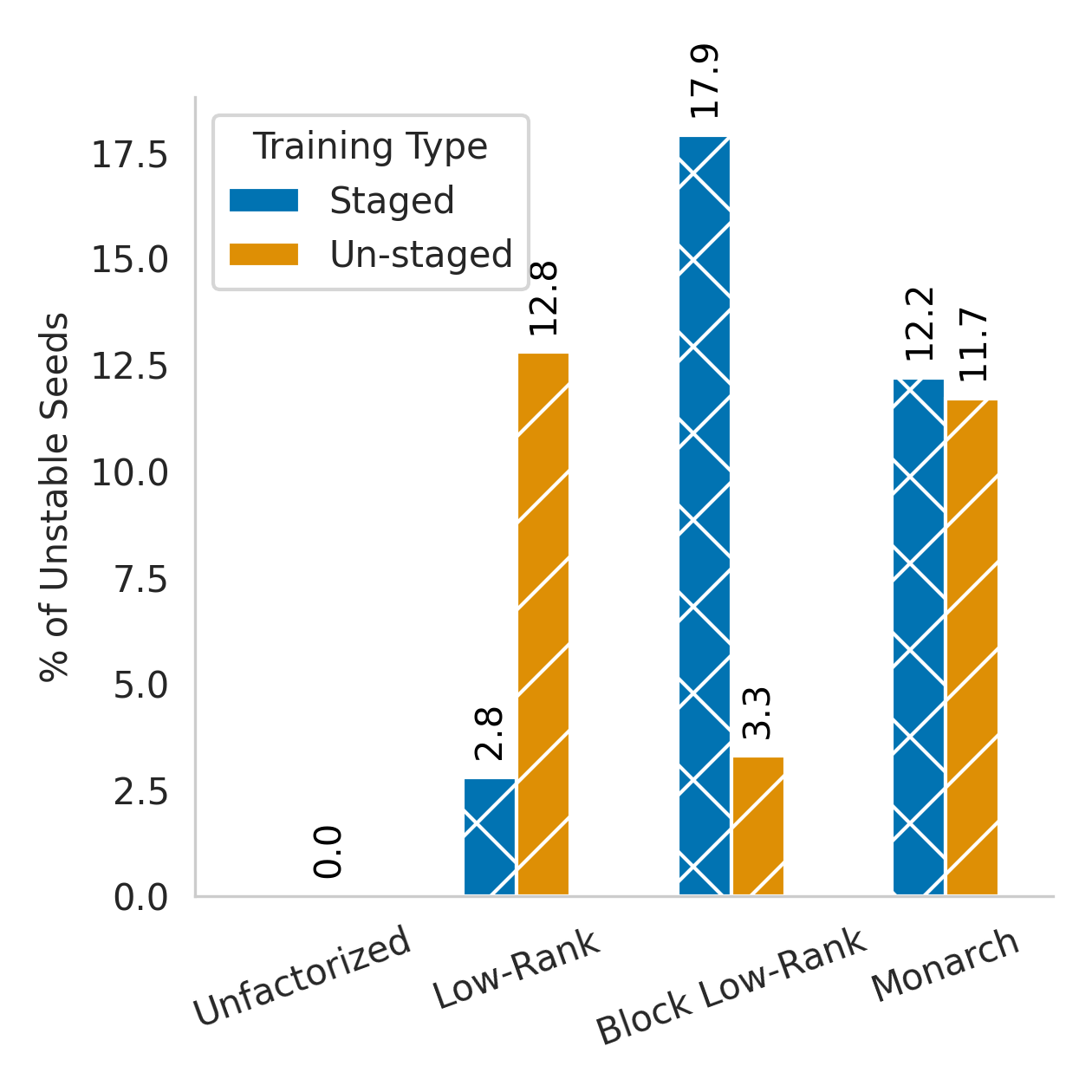}
\caption{Relative stability of factorization methods with and without staging. The results shown here are for compressing the BERT (base) model.
}
\label{fig:stability}
\end{figure}

\begin{figure}
\includegraphics[trim={0mm 0cm 0cm 0cm},clip, width=\linewidth]{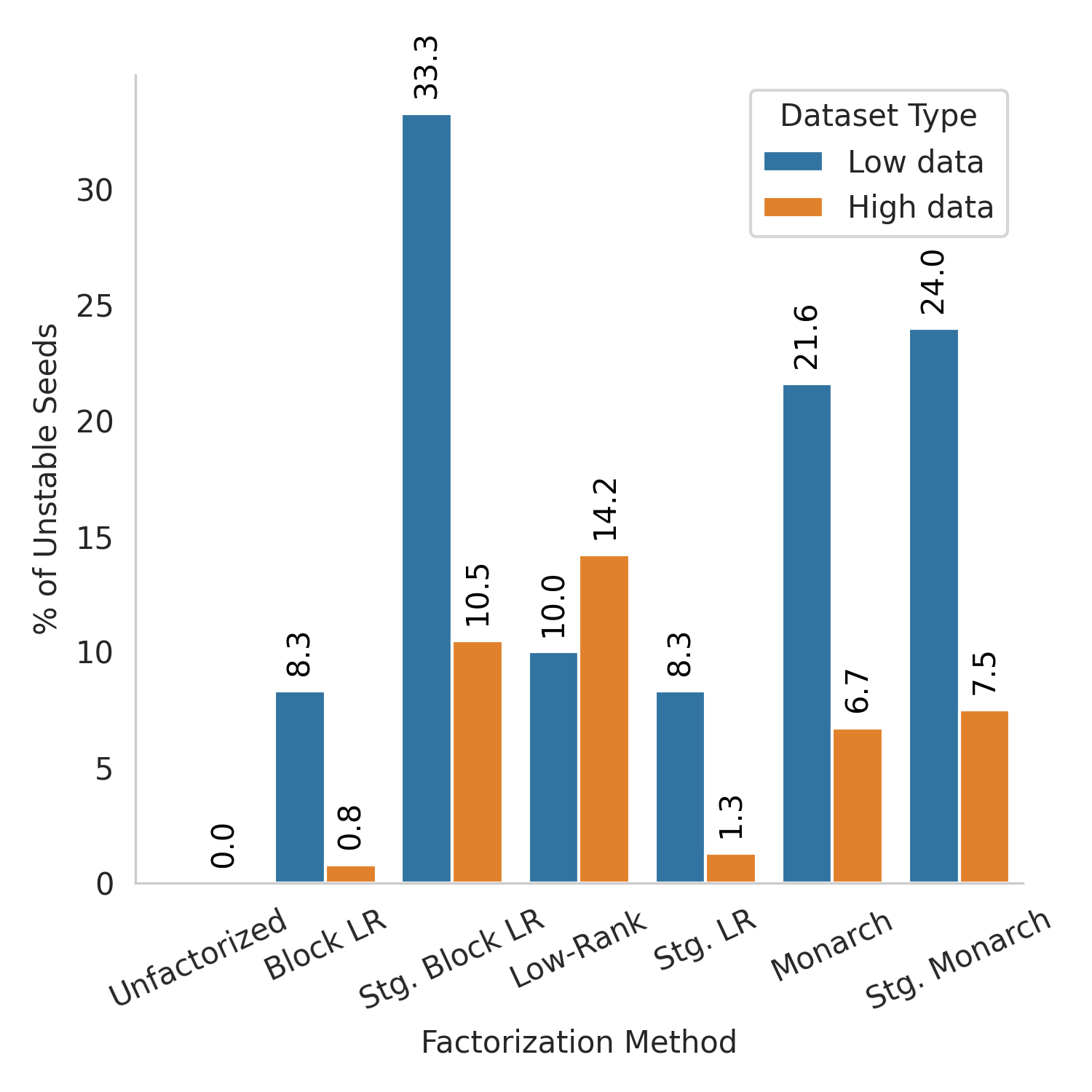}
\caption{
Relative stability of factorization methods across low and high fine-tuning data settings.  
}
\label{fig:stability_dataset}
\end{figure}

\section{Experiments and Results}
\begin{table}[]
\centering
\begin{tabular}{@{}lcc@{}}

\toprule
\textbf{\begin{tabular}[c]{@{}l@{}}Pretrained\\ Model\end{tabular}} & \textbf{\begin{tabular}[c]{@{}l@{}}Params\\ (in M)\end{tabular}} & \textbf{\begin{tabular}[c]{@{}l@{}}Compression\\ Levels (in M)\end{tabular}} \\ \midrule
\multirow{2}{*}{BERT-base}                                 & \multirow{2}{*}{109.5}                                       & 28.5, 44.4                                                          \\
                                                           &                                                              & 56.4, 84.3                                                          \\ \cmidrule(lr){2-3}
\multirow{2}{*}{BERT-large}                                & \multirow{2}{*}{335.1}                                       & 47.3, 89.8                                                          \\
                                                           &                                                              & 160.6, 217.2                                                        \\\cmidrule(lr){2-3}
\multirow{2}{*}{DeBERTa-base}                              & \multirow{2}{*}{184.4}                                       & 103.5, 119.4                                                        \\
                                                           &                                                              & 139.3, 159.2                                                        \\\cmidrule(lr){2-3}
\multirow{2}{*}{T5-Base}                                   & \multirow{2}{*}{222.9}                                       & 34.6, 54.1                                                          \\
                                                           &                                                              & 112.5, 141.7                                                        \\ \bottomrule
\end{tabular}
\caption{Pre-trained models and their compressed variants studied in this work.}
\label{tab:models}
\end{table}
In this section, we will detail the experimental setup for our experiments including the datasets and models used. Following that, we will discuss our results. 
\subsection{Experimental Setup}
\paragraph{Datasets.}
We perform experiments using six classification tasks from the GLUE benchmark~\citep{wang2018glue}. These include the task of textual entailment with MNLI~\citep{williams2018broad} and RTE~\citep{dagan2005pascal}, QNLI~\citep{rajpurkar2016squad}, paraphrase detection with QQP~\footnote{\url{https://quoradata.quora.com/First-Quora-Dataset-Release-Question-Pairs}}, MRPC~\citep{dolan2005automatically}, and sentiment classification with SST-2~\citep{socher2013recursive}. We exclude WNLI~\citep{levesque2012winograd} from our experiments since we found that even the unfactorized model was very unstable. Other details regarding the statistics of these datasets along with metrics used are presented in~\cref{tab:dataset}. Note that GLUE benchmark does not release its test sets and so we use validation sets for evaluation. 

\paragraph{Training Details.} As mentioned earlier, the factorized models exhibit high sensitivity to the choice of the initial learning rate. We, therefore, perform learning rate search with six learning rates $\in \{1e-4, 5e-4, 1e-5, 5e-5, 1e-6, 5e-6\}$. For hyperparameter selection, the models are trained for one epoch and the learning rate that produces the model with the smallest training loss is selected. For all the tasks, we choose 32 as the batch size and use the default 128 as the sequence length. We train all our models using the Transformers library from Huggingface~\citep{wolf2020transformers} with the PyTorch as backend~\citep{paszke2019pytorch}. Additionally, during preliminary experiments, we performed hyperparameter search for the number of training epochs and found five epochs to work well.
Each experiment is performed with six random seeds.~\footnote{We choose seeds from 1 to 6 for reproducibility.}

\begin{table*}[t]
\centering
\begin{tabular}{@{}llccccccc@{}}
\toprule
\multicolumn{2}{c}{\textbf{BERT\textsubscript{Base-Uncased}}} & \multicolumn{7}{c}{\textbf{Task Performance}} \\
\cmidrule(r){1-2} \cmidrule(l){3-9}
\textbf{\# Params (Ratio)} & \textbf{Fact. Method} & \textbf{MNLI} & \textbf{QQP}  & \textbf{SST-2} & \textbf{RTE}  & \textbf{MRPC} & \textbf{QNLI} & \textbf{Avg} \\ 
\midrule
109 M (100\%) &  & 83.6 & 89.4 & 92.9  & 66.8 & 89.4 & 91.0 & 85.5\\
\cmidrule(l){3-9}
\multirow{3}{*}{28.5 M (26\%)}  & \hspace{7pt} Low-Rank & 71.6 & 83.4 & 84.5  & 54.1 & 71.8 & 65.3 & \textbf{71.7} \\
                         & \hspace{7pt} Block LR & 65.6 & 80.0 & 79.7  & 49.1 & 68.1 & 57.4 & 66.6 \\
                         & \hspace{7pt} Monarch  & 66.2 & 75.7 & 78.2  & 49.3 & 74.2 & 58.2 & 66.9 \\
\cmidrule(l){3-9}
\multirow{3}{*}{44.4 M (40\%)}  & \hspace{7pt} Low-Rank & 78.9 & 87.6 & 87.4  & 59.0 & 76.6 & 84.3 & \textbf{78.9} \\
                         & \hspace{7pt} Block LR & 73.1 & 85.5 & 82.8  & 54.0 & 69.8 & 77.4 & 73.7 \\
                         & \hspace{7pt} Monarch  & 70.7 & 83.9 & 81.2  & --   & 70.2 & 58.6 & -- \\
\cmidrule(l){3-9}
\multirow{3}{*}{56.4 M (51\%)}  & \hspace{7pt} Low-Rank & 80.2 & 88.2 & 87.9  & 60.5 & 81.1 & 86.8 & \textbf{80.7} \\
                         & \hspace{7pt} Block LR & 76.6 & 86.4 & 84.6  & 54.7 & 70.2 & 80.8 & 75.5 \\
                         & \hspace{7pt} Monarch  & 74.4 & 85.7 & 82.9  & 52.5 & 70.1 & 79.6 & 74.2 \\
\cmidrule(l){3-9}
\multirow{3}{*}{84.3 M (77\%)}  & \hspace{7pt} Low-Rank & 81.6 & 88.5 & 91.7  & 64.9 & 85.6 & 89.8 & \textbf{83.6} \\
                         & \hspace{7pt} Block LR & 80.5 & 88.3 & 90.0  & 55.5 & 76.9 & 85.2 & 79.4 \\
                         & \hspace{7pt} Monarch  & 79.0 & 87.6 & 86.0  & 53.6 & 72.8 & 80.2 & 76.5 \\ 
\bottomrule
\end{tabular}
\caption{Performance of various factorized versions of BERT\textsubscript{Base-Uncased} on multiple GLUE tasks. Other than the baseline, all values are averaged over successful runs amongst six trials. The right-most column is the average over tasks. Dash indicates that none of the six trials with different seeds were successful in training the model.}
\label{tab:main_bert_base}
\end{table*}

\subsection{Relative Stability of Factorizations}
To study relative stability of different factorizations, we choose BERT's base variant as our test bed. For each of the three factorizations mentioned in~\Cref{sec:factors},  we train them under two configurations: one that uses staged factorization and the other that does not. For all six tasks described in~\cref{tab:dataset}, we train factorized models at 4 different compression levels (shown in~\cref{tab:models}). 

We plot the percentage of unstable seeds across all configurations in~\cref{fig:stability}. We consider a seed to be unstable if it leads to a failed run, i.e., when the classification accuracy is less or equal to that of a majority classifier. We make several key observations regarding the stability of different factorizations. Firstly, it is evident that the unfactorized model exhibits no signs of instability. Secondly, among the factorized models that do not utilize staging, we find that low-rank factorization is the most prone to instability, whereas block low-rank demonstrates the highest stability. Additionally, these results reveal a noteworthy impact of staging on stability: it notably enhances the stability of low-rank factorization, has minimal effect on monarch, and, surprisingly, adversely affects the stability of block low-rank factorization to a significant degree. 

For all the experiments moving forward, we choose staged factorization as our default training strategy for low-rank factorization. For Block Low-Rank, and Monarch, we use their original training strategies that do not use staging.

\paragraph{Impact of Training Data Size on Stability}
~\cref{tab:dataset} highlights a substantial discrepancy in the size of training data across various tasks, with MNLI containing 392k training examples, while RTE contains two orders of magnitude less data, specifically only 2.4K training samples. In light of this variation, we address the question: Does the size of training data correlate with training stability? To explore this relationship, we categorize the six tasks into two groups: those with more than 10k training points are labeled as high-data tasks, while those with fewer are designated as low-data tasks. A clear correlation emerges from the analysis, as depicted in the~\cref{fig:stability_dataset}, revealing that the size of training data is positively correlated with the stability of the factorization. 
A potential explanation for this phenomenon is that the models designed for tasks with limited training data may not receive sufficient training. 
We continued training the models on these datasets beyond their normal duration of five epochs.
However, we observed that increasing the training epochs makes very little to no difference. 

\begin{table*}[]
\centering
\begin{tabular}{@{}llccccccc@{}}
\toprule
\multicolumn{2}{c}{\textbf{T5\textsubscript{Base} Model}} & \multicolumn{7}{c}{\textbf{Task Performance}} \\
\cmidrule(r){1-2} \cmidrule(l){3-9}
\textbf{\# Params (Ratio)} & \textbf{Fact. Method} & \textbf{MNLI} & \textbf{QQP}  & \textbf{SST-2} & \textbf{RTE}  & \textbf{MRPC} & \textbf{QNLI} & \textbf{Avg} \\ 
\midrule
222.9 M (100\%) &  & 86.6 & 90.1 & 94.5  & 80.1 & 91.2 & 93.1 & 87.0\\
\cmidrule(l){3-9}
\multirow{3}{*}{34.6 M (15\%)}   & \hspace{7pt} Low-Rank & 75.4 & 81.7 & 86.6 & 52.3 & 75.4 & 69.7 & 73.5 \\
                                 & \hspace{7pt} Block LR & 71.1 & 83.6 & 85.5 & 52.6 & 75.5 & 68.8 & 72.8 \\
                                 & \hspace{7pt} Monarch  & 75.3 & 85.8 & 86.7 & 52.6 & 75.6 & 66.8 & 73.8 \\
\cmidrule(l){3-9}
\multirow{3}{*}{54.1 M (24\%)}   & \hspace{7pt} Low-Rank & 80.0 & 86.8 & 88.5 & 58.6 & 80.6 & 86.1 & 80.1 \\
                                 & \hspace{7pt} Block LR & 74.4 & 84.7 & 85.9 & 52.4 & 73.7 & 68.9 & 73.3 \\
                                 & \hspace{7pt} Monarch  & 77.9 & 87.7 & 87.9 & 52.1 & 75.9 & 83.9 & 77.5 \\
\cmidrule(l){3-9}
\multirow{3}{*}{112.5 M (50\%)}  & \hspace{7pt} Low-Rank & 84.0 & 89.3 & 90.9 & 74.4 & 87.7 & 90.0 & 86.0 \\
                                 & \hspace{7pt} Block LR & 71.6 & 85.2 & 86.3 & 50.9 & 74.6 & 68.9 & 72.9 \\
                                 & \hspace{7pt} Monarch  & 81.2 & 88.8 & 89.0 & 55.5 & 83.8 & 85.9 & 80.7 \\
\cmidrule(l){3-9}
\multirow{3}{*}{141.7 M (63\%)}  & \hspace{7pt} Low-Rank & 84.9 & 89.6 & 91.2 & 77.1 & 88.0 & 90.9 & 86.9 \\
                                 & \hspace{7pt} Block LR & 77.6 & 84.6 & 84.7 & 52.2 & 74.6 & 69.0 & 73.7 \\
                                 & \hspace{7pt} Monarch  & 84.5 & 89.3 & 90.1 & 61.1 & 85.9 & 87.3 & 83.0 \\ 
\bottomrule
\end{tabular}
\caption{Performance of various factorized versions of T5\textsubscript{Base} on multiple GLUE tasks. Other than the baseline, all values are averaged over successful runs amongst six trials. The right-most column is the average over tasks. Dash indicates that none of the six trials with different seeds were successful in training the model.}
\label{tab:t5-base}
\end{table*}

\subsection{Which Factorization is Most Effective?}
\paragraph{Pre-trained Models.} We compare the three factorization methods on four pre-trained models specified in~\cref{tab:models}. 
Among these, BERT~\citep{devlin2019bert} and DeBERTa~\citep{he2020deberta} are encoder-only models that are trained with masked language modeling. We also perform experiments with a T5 model which has an encoder-decoder architecture. 
This allows us to investigate whether any observations remain consistent across different types of architectures.
We do not consider any decoder-only models such as GPT-2~\citep{radford2019language} or OPT~\citep{zhang2022opt}, since they are primarily used for generative tasks and severly underperform on classification tasks. 

\paragraph{Compression Levels.} Each of the factorization method has knobs that control for the number of parameters. For each of the pre-trained models, we factorize models at four difference compression levels (shown in ~\cref{tab:models}).

\paragraph{Results.} 
The comparison of results for the three factorization methods on six tasks is presented in tables~\cref{tab:main_bert_base} and ~\cref{tab:t5-base}. 
The reported scores in each cell are an average of stable seeds for that model, i.e., we ignore the unstable seeds while calculating the average score.
Across all pre-trained models, the low-rank factorization consistently demonstrates superior performance compared to the other two factorization methods.  Specifically, on the base variant of BERT, low-rank factorization consistently outperforms the other two by an average of 4-5\% in scores. Notably, this superiority is maintained consistently across various compression levels. 
 Next, we compare results for the T5 model, see~\cref{tab:}. While low-rank factorization still performs the best, the block low-rank factorization performs significantly worse than the Monarch factorization.

Note that while Block Low-Rank and Monarch implementations are more complex (requiring einsums), the implementation for low-rank factorization is the simplest --it is simply an application of two linear layers one after the other.

\paragraph{Local vs Global Redundancies.} The primary rationale behind incorporating Block Low-Rank factorization is to investigate whether the local redundancies utilized through blockwise factorization offer greater benefits than those exploited through matrix-level factorization. The results of our study, however, demonstrate that Block Low-Rank factorization performs less effectively compared to low-rank factorization at the matrix level.  

\paragraph{Block Low-Rank vs Monarch.} 
Another reason for introducing Block Low-Rank factorization was to conduct a comparison with Monarch. The key distinction lies in the permutation operation implemented by Monarch, which rearranges the elements in the original matrix before executing low-rank factorization at the block level. Our findings indicate that the effectiveness of either factorization depends on the pre-trained model—Block Low-Rank surpasses Monarch for BERT models (both base and large), whereas the reverse is observed for the T5 model. 
While not definitive, this observation is logical --which elements in a parameter block harbor redundancy can be influenced by numerous factors, such as training data, model architecture, and a myriad of other choices made during training.

\begin{table}[]
\centering
\begin{tabular}{@{}lc@{}}
\toprule
\textbf{Factorization}  & \textbf{\begin{tabular}[c]{@{}c@{}}\% Unstable\\  Seeds\end{tabular}} \\ \midrule
Unfactorized       & 0.0 \\
Low-Rank       & 0.0                                                          \\
Block Low-Rank & 0.0                                                          \\
Monarch        & 1.1                                                          \\ \bottomrule
\end{tabular}
\caption{Relative Instability for T5 model. All factorizations are much more stable for the T5 model. }
\label{tab:t5_stability}
\end{table}

\paragraph{Stability on T5 Factorized Models.} Among all models, the models factorized and fine-tuned using T5 (base) are the most stable. We show the relative instability of three factorization for the T5 model in~\cref{tab:t5_stability}. This is in stark contrast to the results observed for BERT-base where even the most stable factorization (Low-Rank with Staging) was twice as unstable as Monarch. One potential consequence of these results is that T5 models are more amenable to matrix factorizations and can be prioritized over the other encoder-only models. We hypothesize that the increased stability of the T5 model can be attributed to the fact that pre-training for T5 includes the classification datasets from the GLUE benchmark~\citep{raffel2020exploring}.

\paragraph{BERT-base vs BERT-large.}
We note an intriguing finding in the comparison between BERT-base and BERT-large models~\cref{tab:base-large}. Presenting two configurations of compressed models with parameters within a 10\% range of each other, it is noteworthy that the model compressed from BERT-base (44.5M) demonstrates a significant performance advantage over its counterpart compressed from BERT-large (47.3M). It's worth noting that a compressed BERT-base with 44.5 million parameters performs favorably when compared to a model twice its size (compressed from BERT-large). This deviation from conventional expectations, which hinge on model size alone, implies that factors beyond the sheer count of parameters significantly influence model performance.

\begin{table}[]
\centering
\resizebox{0.98\columnwidth}{!}{
\begin{tabular}{@{}llllll@{}}
\toprule
\textbf{Model}        & \textbf{MNLI} & \textbf{QQP}  & \textbf{QNLI} & \textbf{SST-2} & \textbf{Avg.} \\ \midrule
Base - 44.5  & 78.9 & 87.6 & 84.3 & 87.4  & \textbf{84.6 }\\
Large - 47.3 & 70.1 & 84.6 & 76.5 & 84.4  & 78.9 \\
Base - 84.3  & 81.6 & 88.5 & 89.8 & 91.7  & \textbf{87.9} \\
Large - 89.8 & 79.4 & 87.7 & 85.6 & 88.2  & 85.2 \\ \bottomrule
\end{tabular}
}
\caption{Smaller compressed models outperform larger ones. BERT-base vs BERT-large.}
\label{tab:base-large}
\end{table}

\begin{table}[]
\centering
\begin{tabular}{@{}lrr@{}}
\toprule
\multirow{2}{*}{\textbf{Model}} & \multicolumn{2}{c}{\textbf{\begin{tabular}[c]{@{}c@{}}Inference\\ Time (ms)\end{tabular}}} \\ \cmidrule(l){2-3} 
                                & \textbf{28.5 M}                              & \textbf{84.3 M}                             \\ \midrule
BERT (full)                     & 8.87                                         & 8.87                                        \\
Monarch                         & 5.09                                         & 14.82                                       \\
Block Low-Rank                  & 4.72                                         & 14.42                                            \\
Low-Rank                        & 3.10                                         & 7.29                                        \\ \bottomrule
\end{tabular}
\caption{Inference times for the Factorized models.}
\label{tab:inference}
\end{table}
\paragraph{Compute Efficiency.} We report inference times for the three factorization methods along with the full model in~\cref{tab:inference}. We use the \texttt{torch.cuda.Event} module from PyTorch to calculate the average inference time (forward pass).~\footnote{We account for GPU warm-up and omit first two runs from the average.} The results are averaged over 8 runs with 100 random examples.  Not only is implementing low-rank factorization straightforward, we observe that it is also computationally more efficient. With the largest factorized model, while low-rank factorization has lower latency than the unfactorized model, the inference time for Block Low-Rank and Monarch are significantly higher.

A caveat is important to note --we could not get access to the Monarch implementation for BERT and therefore the results here are from our own implementation using the \texttt{einsum} functionality available in PyTorch.~\footnote{The official implementation at\url{https://github.com/HazyResearch/fly} does not include implementation for BERT and rectangular Monarch matrices.}

\section{Conclusion}
In conclusion, our study highlights the critical importance of addressing stability issues in factorized transformer models, especially those associated with low-rank factorization. The introduction of the Staged Factorization training strategy proves to be a promising solution, substantially improving the stability of low-rank factorization and aligning it closely with the stability observed in unfactorized models. Furthermore, our extensive experiments on diverse classification datasets affirm the dominance of straightforward Low-Rank Factorization over the recently proposed Monarch factorization. Not only does it achieve higher accuracy, but also requires the lowest inference time, establishing itself as a superior choice for factorized transformer models. The low-rank factorization is easy-to-implement and is many times faster than the more sophisticated factorization methods like Monarch. 
\section{Limitations}
Our study acknowledges certain limitations that warrant consideration. Firstly, we recognize that the emergence of few-shot in-context models represents a significant evolution in transformer architectures. However, our exploration is confined to traditional models, and we do not extend our factorization analysis to these newer variants. 

Secondly, our experiments exclusively focus on matrix factorization methods, and we refrain from investigating tensor factorizations such as Tensor-Train, Tucker, and others. This limitation arises, in part, from the absence of standardized approximation procedures for these tensor factorization methods. The lack of a universally accepted approach makes it challenging to conduct a comprehensive and fair comparison with matrix factorization techniques. While our study provides valuable insights into matrix factorization, it leaves an avenue open for future research to explore the implications and performance of tensor factorization methods in the context of transformer models.
\bibliography{custom}

\appendix
\section{Additional Experiments and Information}
\begin{figure*}[t]
\centering
\includegraphics[trim={1.2cm 0cm 0cm 1.2cm},clip, width=.48\textwidth]{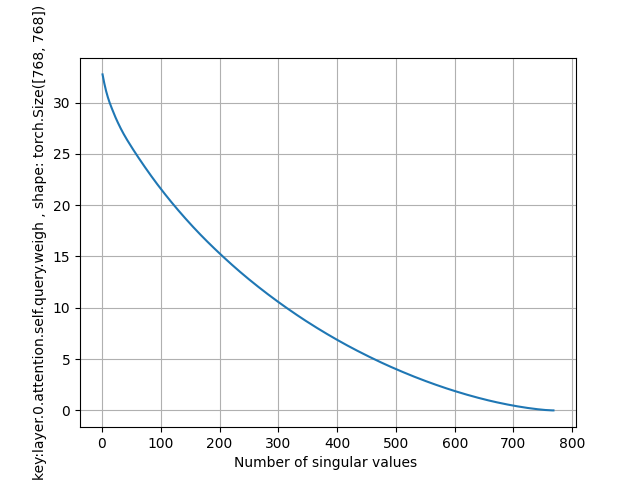}
\includegraphics[trim={1.2cm 0cm 0cm 1.2cm},clip, width=.48\textwidth]{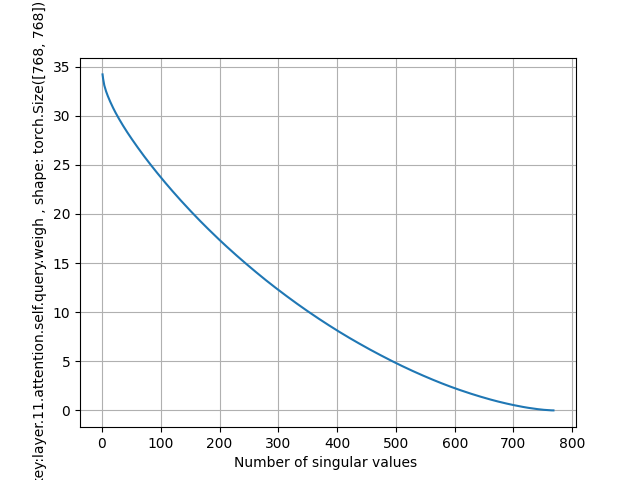}
\caption{\textbf{Reconstruction Error vs No. of Singular Values.} The matrix on the left is the query matrix from the first layer of the pre-trained BERT model(~\texttt{bert-base-uncased}), and the right is for the query matrix of the last layer (12th). As can been seen, pre-trained matrices in these pre-trained models are close to full-rank. 
}
\label{fig:spectrum}
\end{figure*}

\begin{figure*}[]
\centering
\includegraphics[trim={0cm 0cm 0cm 0cm},clip]{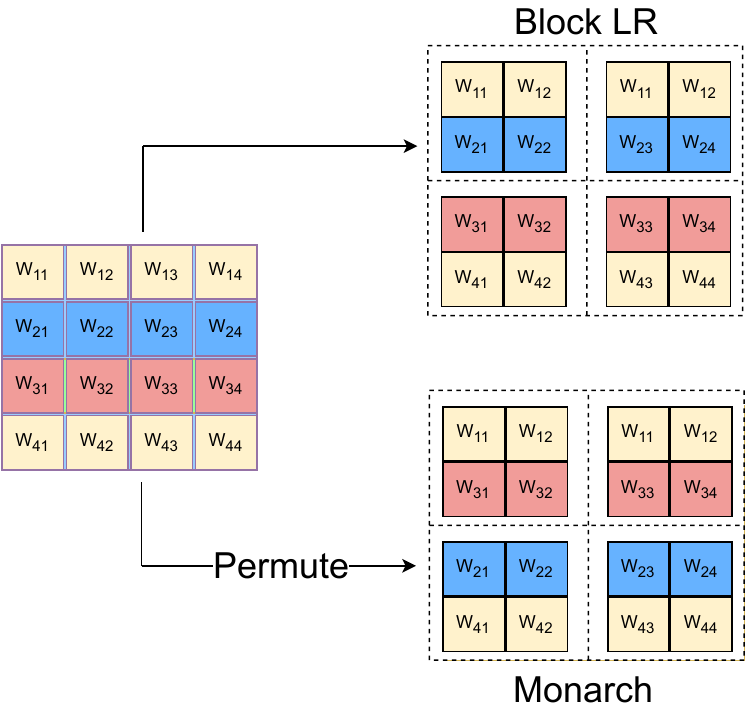}
\caption{An example of a $4 x 4$ matrix for Block Low-Rank and Monarch Factorizations. Monarch Factorization operates on the permutation of the elements of the original matrix.
}
\label{fig:example}
\end{figure*}

\subsection{Computing Infrastructure Used}
All of our experiments required access to GPU accelerators. We ran our experiments on three machines: Nvidia Tesla A100 (80 GB VRAM), Nvidia Tesla V100 (16 GB VRAM), Tesla A100 (40 GB VRAM). 


We include the performance of various factorization methods for BERT\textsubscript{Large-Uncased} and DeBERTa models as well.
\begin{table*}[]
\centering
\begin{tabular}{@{}llccccccc@{}}
\toprule
\multicolumn{2}{c}{\textbf{BERT\textsubscript{Large-Uncased} Model}} & \multicolumn{7}{c}{\textbf{Task Performance}} \\
\cmidrule(r){1-2} \cmidrule(l){3-9}
\textbf{\# Params (Ratio)} & \textbf{Fact. Method} & \textbf{MNLI} & \textbf{QQP}  & \textbf{SST-2} & \textbf{RTE}  & \textbf{MRPC} & \textbf{QNLI} & \textbf{Avg} \\ 
\midrule
335.1 M (100\%) &  & 85.7 & 89.8 & 93.9  & 70.8 & 90.3 & 91.8 & 87.0\\
\cmidrule(l){3-9}
\multirow{3}{*}{47.3 M (14\%)}  & \hspace{7pt} Low-Rank & 70.1    & 84.6 & 84.4  & 54.8 & --    & 76.5 & - \\
                                & \hspace{7pt} Block LR & 64.6 & 80.7 & 79.4  & 50.4 & 70.2 & 58.1 & 67.2 \\
                                & \hspace{7pt} Monarch  & 64.1 & 81.1 & 79.5  & --    & 73.8 & 57.5 & -- \\
\cmidrule(l){3-9}
\multirow{3}{*}{89.8 M (26\%)}  & \hspace{7pt} Low-Rank & 79.4 & 87.5 & 88.2  & 56.5 & --    & 85.6 & -- \\
                                & \hspace{7pt} Block LR & 72.1 & 85.2 & 79.8  & 52.4 & 70.7 & 76.2 & 72.7 \\
                                & \hspace{7pt} Monarch  & 70.1 & 84.3 & 80.3  & 49.3 & 72.2 & 61.2 & 69.5 \\
\cmidrule(l){3-9}
\multirow{3}{*}{160.6 M (47\%)}  & \hspace{7pt} Low-Rank & 83.6 & 89.2 & 90.8  & 67.2 & 86.3 & 90.1 & 84.5 \\
                                 & \hspace{7pt} Block LR & 76.2 & 86.5 & 83.4  & 54.9 & 73.5 & 83.0 & 76.2 \\
                                 & \hspace{7pt} Monarch  & 69.8 & 84.7 & 81.2  & 55.2 & --    & 75.5 & -- \\
\cmidrule(l){3-9}
\multirow{3}{*}{217.2 M (64\%)}  & \hspace{7pt} Low-Rank & 84.6 & 89.6 & 92.1  & 69.9 & 85.5 & 90.9 & 85.4 \\
                                 & \hspace{7pt} Block LR & 78.0   & 86.7 & 87.6  & 51.5 & 74.4 & 85.8 & -- \\
                                 & \hspace{7pt} Monarch  & 75.4 & 86.4 & --     & 50.9 & 74.4 & 83.6 & -- \\ 
\bottomrule
\end{tabular}
\caption{Performance of various factorized versions of BERT\textsubscript{Large-Uncased} on multiple GLUE tasks. Other than the baseline, all values are averaged over successful runs amongst six trials. The right-most column is the average over tasks. Dash indicates that none of the six trials with different seeds were successful in training the model.}
\end{table*}

\begin{table*}[]
\centering
\begin{tabular}{@{}llccccccc@{}}
\toprule
\multicolumn{2}{c}{\textbf{DeBERTa Model}} & \multicolumn{7}{c}{\textbf{Task Performance}} \\
\cmidrule(r){1-2} \cmidrule(l){3-9}
\textbf{\# Params (Ratio)} & \textbf{Fact. Method} & \textbf{MNLI} & \textbf{QQP}  & \textbf{SST-2} & \textbf{RTE}  & \textbf{MRPC} & \textbf{QNLI} & \textbf{Avg} \\ 
\midrule
184.4 M (100\%) &  & 89.8 & 91.1 & 94.9  & 82.2 & 91.9 & 93.9 & 87.0\\
\cmidrule(l){3-9}
\multirow{3}{*}{103.5 M (56\%)}  & \hspace{7pt} Low-Rank & 76.4 & 84.6 & 87.0  & 51.3 & --    & 81.2 & -- \\
                                 & \hspace{7pt} Block LR & 39.6 & 82.1 & 83.5  & 54.5 & 73.8 & 59.9 & 65.5 \\
                                 & \hspace{7pt} Monarch  & --    & 70.7 & --     & --    & --    & --    & -- \\
\cmidrule(l){3-9}
\multirow{3}{*}{119.4 M (64\%)}  & \hspace{7pt} Low-Rank & 82.4 & 88.4 & 88.9  & 62.8 & 82.8 & 87.4 & 82.1 \\
                                 & \hspace{7pt} Block LR & 70.5 & 85.3 & 84.6  & --    & --    & 80.1 & --    \\
                                 & \hspace{7pt} Monarch  & --   & --    & --     & 52.2 & 50.0 & 58.1 & --    \\
\cmidrule(l){3-9}
\multirow{3}{*}{139.3 M (75\%)}  & \hspace{7pt} Low-Rank & 84.5 & 88.9 & 92.1  & 70.9 & 86.7 & 89.6 & 85.4 \\
                                 & \hspace{7pt} Block LR & 78.8 & 87.5 & 86.0  & 51.1 & 75.1 & 82.0 & 76.7 \\
                                 & \hspace{7pt} Monarch  & 55.7 & 75.9 & 79.8  & --    & --    & --    & -- \\
\cmidrule(l){3-9}
\multirow{3}{*}{159.2 M (86\%)}  & \hspace{7pt} Low-Rank & 88.1 & 90.0 & 93.5  & 75.3 & 87.9 & 92.1 & 87.8 \\
                                 & \hspace{7pt} Block LR & 84.3 & 88.7 & 89.2  & 57.1 & 84.0 & 84.0 & -- \\
                                 & \hspace{7pt} Monarch  & 68.6 & 83.6 & --     & --    & --    & 63.7 & -- \\ 
\bottomrule
\end{tabular}
\caption{Performance of various factorized versions of DeBERTa on multiple GLUE tasks. Other than the baseline, all values are averaged over successful runs amongst six trials. The right-most column is the average over tasks. Dash indicates that none of the six trials with different seeds were successful in training the model.}
\end{table*}



\end{document}